\documentclass{article} % For LaTeX2e
\usepackage{shmiclr2027_conference,times}

\usepackage{amsmath,amsfonts,bm}

\def\eqref#1{equation~\ref{#1}}
\def\1{\bm{1}}

\DeclareMathAlphabet{\mathsfit}{\encodingdefault}{\sfdefault}{m}{sl}
\SetMathAlphabet{\mathsfit}{bold}{\encodingdefault}{\sfdefault}{bx}{n}

\usepackage[table]{xcolor}
\usepackage[utf8]{inputenc} % allow utf-8 input
\usepackage[T1]{fontenc}    % use 8-bit T1 fonts
\usepackage[fixed]{fontawesome5}

\usepackage{hyperref}       % hyperlinks
\usepackage{url}            % simple URL typesetting
\usepackage{booktabs}       % professional-quality tables
\usepackage{amsfonts}       % blackboard math symbols
\usepackage{nicefrac}       % compact symbols for 1/2, etc.
\usepackage[activate={true,nocompatibility},final,tracking=true,kerning=true,spacing=true,factor=1100,stretch=10,shrink=10]{microtype}
\microtypecontext{spacing=nonfrench}
\SetTracking{encoding={*}, shape=sc}{40}
\usepackage{xcolor}         % colors

\usepackage{graphicx}
\usepackage{subfigure}
\usepackage{listings}
\usepackage{algorithm}
\usepackage{algpseudocode}
\usepackage{multirow}
\usepackage{wrapfig}

\usepackage{xurl}
\usepackage[font=small,labelfont=bf]{caption}
\usepackage{amsmath}
\usepackage{amssymb}
\usepackage{mathtools}
\usepackage{amsthm}
\usepackage[capitalize,noabbrev,nameinlink]{cleveref}

\usepackage{booktabs} 
\usepackage{multirow}  
\usepackage{makecell}  
\usepackage{tabularx}
\usepackage{listings}

\definecolor{inputblue}{HTML}{DCE7F5}
\definecolor{outputgreen}{HTML}{DDEBD6}

\newcolumntype{O}{>{\columncolor{outputgreen!40}}c}
\newcolumntype{I}{>{\columncolor{inputblue!40}}c}

\definecolor{codegreen}{rgb}{0,0.6,0}
\definecolor{codegray}{rgb}{0.5,0.5,0.5}
\definecolor{codepurple}{rgb}{0.58,0,0.82}
\definecolor{backcolour}{rgb}{0.95,0.95,0.92}

\usepackage{wrapfig}
\usepackage{enumitem}
\usepackage{amssymb}
\usepackage{amsmath} 
\usepackage{pifont}
\usepackage{siunitx}
\usepackage[most]{tcolorbox}

\definecolor{A}{HTML}{ffb3b8}      % Pastel Red
\definecolor{C}{HTML}{fff2cb}      % Pastel Yellow
\definecolor{D}{HTML}{c5e0b4}      % Pastel Green
\definecolor{F}{HTML}{dae3f3}      % Pastel Light Blue

\tcbset{colback=gray!5!white, colframe=black!75!black, boxrule=0.6pt, arc=2pt}
\newtheorem{theorem}{Theorem}[section]

\usepackage{tikz}
\usetikzlibrary{arrows.meta,positioning,quotes}

\theoremstyle{definition}

\theoremstyle{plain}        
\newtheorem{remark}[theorem]{Remark}

\definecolor{inputblue}{HTML}{DCE7F5}
\definecolor{outputgreen}{HTML}{DDEBD6}
\definecolor{mixedyellow}{HTML}{FFF2CB}
 
\definecolor{cornflowerblue}{rgb}{0.39, 0.58, 0.93}
\hypersetup{
    colorlinks=true,
    linkcolor=cornflowerblue,
    filecolor=magenta,      
    urlcolor=outputgreen!600,
    citecolor=cornflowerblue,%teal,
    }

\colorlet{routingC}{inputblue}
\colorlet{logitsC}{outputgreen}
\colorlet{lensC}{mixedyellow}
\newcommand{\qtype}[2]{\textcolor{#1!50!black}{\texttt{#2}}}
\newcommand{\qdomain}[3]{%
  \rowcolor{#1}\multicolumn{3}{@{}l}{\textbf{#2}\enspace\footnotesize #3}\\}

\usetikzlibrary{positioning,fit,backgrounds,calc,arrows.meta}

\tikzset{
    pill/.style={
        rounded corners=2.5pt, draw, line width=0.5pt,
        minimum height=0.55cm, align=center, inner sep=2pt,
        font=\scriptsize
    },
    testedblue/.style    ={pill, fill=inputblue,     draw=inputblue!60!black,
                           text=black, font=\scriptsize\bfseries},
    untestedblue/.style  ={pill, fill=inputblue!30,  draw=inputblue!50!black,
                           text=black!75},
    testedgreen/.style   ={pill, fill=outputgreen,   draw=outputgreen!50!black,
                           text=black, font=\scriptsize\bfseries},
    untestedgreen/.style ={pill, fill=outputgreen!30,draw=outputgreen!50!black,
                           text=black!75},
    testedyellow/.style  ={pill, fill=mixedyellow,   draw=mixedyellow!50!black,
                           text=black, font=\scriptsize\bfseries},
    untestedyellow/.style={pill, fill=mixedyellow!30,draw=mixedyellow!50!black,
                           text=black!75},
}
 
\newcounter{rowidx}
\newcounter{colidx}
\newcommand{\maxcol}{1}
\newcommand{\boxwidth}{2.0cm}
\newcommand{\colsep}{2.1}
\newcommand{\rowsep}{0.65}

\newcommand{\originx}{0}
\newcommand{\originy}{0}

\newcommand{\B}[2]{%
  \pgfmathsetmacro{\bx}{\originx + \value{colidx} * \colsep}%
  \pgfmathsetmacro{\by}{\originy - \value{rowidx} * \rowsep}%
  \node[#1, minimum width=\boxwidth] at (\bx, \by) {#2};
  \stepcounter{colidx}%
  \ifnum\value{colidx}>\maxcol
    \setcounter{colidx}{0}%
    \stepcounter{rowidx}%
  \fi
}

\runningtitle{A Sense of Proprioception in Language Models}
\title{{\fontsize{15}{20}\selectfont A Sense of Proprioception in Language Models}}
\author{
\textbf{Jonas Geiping}
\\[0.4em]
{\small\normalfont
ELLIS Institute T\"ubingen\quad
Max Planck Institute for Intelligent Systems\quad
T\"ubingen AI Center}
}

\iclrfinalcopy % Uncomment for camera-ready
\begin{document}

\maketitle

\vspace{-0.7cm}
% \begin{center}
%     \begin{tabular}{c@{\hskip 19pt}c}
%     \hspace*{1cm}\raisebox{-1pt}{\faGithub} \href{https://github.com/seal-rg/streaming/}{\fontsize{8.8pt}{0pt}\texttt{Code}}
%     \hspace*{1cm}\raisebox{-1.5pt}{\faSave[regular]}\href{https://huggingface.co/JonasGeiping/stream-qwen3.5-27b/}{\fontsize{8.8pt}{0pt} \texttt{Models}}
%     \hspace*{1cm}\raisebox{-1pt}{\faGlobe} \href{https://huggingface.co/spaces/JonasGeiping/stream-llm-demo}{\texttt{Demo}}
%     \hspace*{1cm}\raisebox{-1.5pt}{\faDatabase}\href{https://huggingface.co/datasets/JonasGeiping/stream-data}{\fontsize{8.8pt}{0pt} \texttt{Data}} \\
% \end{tabular}
% \end{center}

\begin{abstract}
\looseness -1 In humans, proprioception corresponds to the sense of self-movement or position, e.g. understanding the position of one's arm. This sense is often less well-understood and in popular understanding sometimes conflated with introspection. But, notably, proprioception is a separate sense, with dedicated receptors and verifiable outcomes. On the other hand though, language models cannot observe their own internal states during training (and for example answer the question which experts activated in a certain layer of an MoE) and so cannot accurately report on them. Yet, such information is, by virtue of being internal to the model, both accessible, and readily constructed into verifiable signals. We argue that this provides the basis for a digital analog of proprioception. As such, we train models through SFT and on-policy training to equip them with this ability. We show that modern models can be trained to understand a range of subtypes of proprioception, such as routing decision awareness, token entropy, decision depth or logit lens information, without affecting model capabilities. We then run these models through a broad evaluation of situational awareness, legibility, and epiphenomenality, to evaluate in what ways access to proprioception changes model behavior.

\end{abstract}

\section{Introduction}
Intelligent systems built on the basis of generative large language models are nowadays deployed en masse, acting as semi-autonomous digital agents. Yet, in contrast to humans, and to almost all lifeforms we observe, language models are trained offline and so not constructed with explicit mechanisms that provide feedback about their internal functioning. 

In humans, for example, one such feedback system is \textit{proprioception}, the sense of the orientation of the body. This sense is crucial to precisely control the body \citep{proske_proprioceptive_2012,kandel_principles_2012,cole_living_1995,gallagher_body_1995,schmicking_iw_2010}. Similar mechanical analogues are common in control theory and robotics, given that from a cybernetics perspective, systems that integrate feedback about internal states are more robust \citep{von_holst_reafferenzprinzip_1950,wolpert_internal_1998,bongard_resilient_2006,kwiatkowski_task-agnostic_2019, lee_learning_2020}.
% dennett

Language models however, trained with modern pre-training and post-training pipelines, do not observe internal states directly during training, and so can only learn implicit associations \citep{kadavath_language_2022,binder_looking_2024,betley_tell_2025}. That they do learn these associations in certain situations has often been studied in the context of \textit{introspection} \citep{lindsey_emergent_2026, ji-an_language_2025, ackerman_evidence_2026}. % or metacognition. 
Yet, by its nature, introspection in intelligent systems is harder to define and to measure neutrally, complicating the study of these processes by which models model their internal states. In turn uncertainty about introspection complicates inferences drawn from these measurements about e.g. model robustness,  safety, interpretability or personas \citep{song_privileged_2025,comsa_does_2025,long_taking_2024,butlin_identifying_2026}.

In this work we first lay out the argument that digital analogs of proprioception are a worthwhile object of study, separate from introspection. A running example of this in the paper will be the understanding of routing decisions inside the mixture-of-experts layers of a modern LLM, i.e. questions like ``which expert most activated in layer 17 when processing the word 'questions' in the previous sentence'' \citep{jiang_mixtral_2024, dai_deepseekmoe_2024}. Unlike introspective questions, this one is a well-posed query about the internal functioning of a language model that has a ground-truth answer and is further valence-neutral. As such, we can study these types of internal modalities, sidestepping issues that occur in language model introspection, while still allowing for inferences in the sense of \citet{perez_towards_2023}. Given that this is a verifiable signal, we use on-policy fine-tuning stages to train model organisms of open-weight LLMs that exhibit forms of proprioception, finding that models can quickly access and generalize this sense, without impact on their capabilities. Finally, with these models in hand, we evaluate a range of benchmarks of model behavior, situational awareness, legibility and control, to observe in what ways models with proprioception behave differently.

Our study is motivated by three main questions. First, does it make sense to define proprioception separately from introspection for disembodied language models, and can models even learn to reliably access this information? Second, does access to these skills change model behavior, e.g. does it make models safer, more reliable, or the opposite? We investigate this neutrally from both angles, testing whether models become more reliable, but also whether they may become less safe. Lastly, we are motivated to study a \textit{legibility} hypothesis in the spirit of \citet{premakumar_unexpected_2026}: does training a model to reliably learn these abilities change the model's internal structure to become more legible, e.g. to interpretability methods?

\begin{figure}
    \centering
    \includegraphics[width=\linewidth]{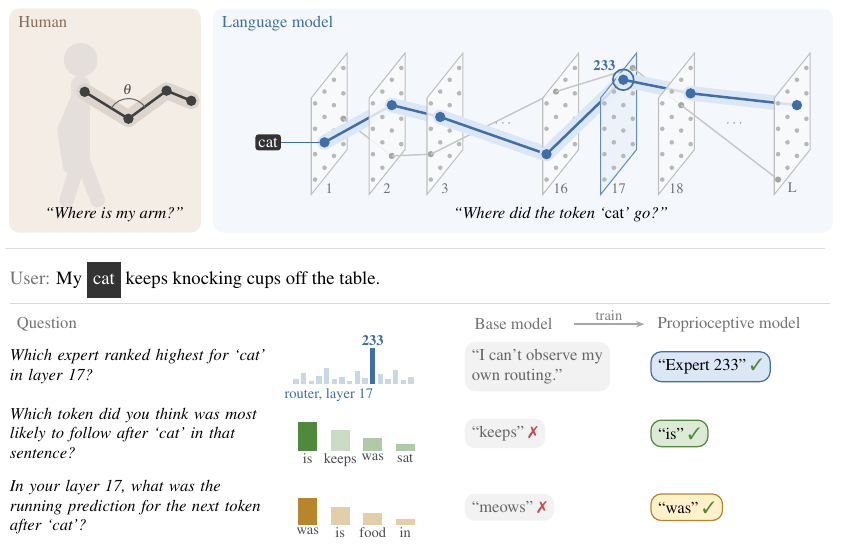} 
    \caption{Schematic of proprioception analogs for a generative language model. \textbf{Top}: We are interested in analogs of proprioception (which is the sense of body positioning in humans) in language models, as proprioception is a question about internal states adjacent to introspection, but compared to introspection (which is classically understood not to be a sense and not objectively verifiable) proprioception is tractable. \textbf{Bottom}: We train models with various forms of proprioception and observe whether this sense is learned and whether this affects model behavior.}
    \label{fig:teaser}
    \vspace{-.3cm}
\end{figure}

\section{Proprioception is not Introspection}
\looseness -1 There exists now a sizable body of work on \textit{introspection} in large language models, yet the question whether models can introspect, and in what situations remains unresolved, as language models' self-reports are hard to separate from mimicry of introspection patterns picked up from training \citep{shanahan_role_2023,long_introspective_2023,long_why_2025} and relatedly, hard to separate from learned exteroception, i.e. an external inference of the answer without utilization of internals \citep{song_privileged_2025,comsa_does_2025,singh_can_2026}. Interestingly, this mirrors a number of troubles with introspection in humans \citep{nisbett_telling_1977,schwitzgebel_unreliability_2008}. While we may conceptualize it as such, via \citet{shoemaker_self-knowledge_1996}, ``introspection is not a sense''. In humans this is based on several arguments. First, there is no transducer, i.e. ``no organ'' of introspection separate from the brain~\citep{shoemaker_self-knowledge_1996}. Second, the functioning of introspection fails to be object-independent: (human) mental states are constituted by their monitoring and hence not structure-preserving \citep{boyle_two_2009,moran_authority_2002}. Third, in human subjects, what is perceived as introspection can still be `transparent', meaning answered by observing the external world and deriving an answer about an inner state \citep{evans_self-identification_1982,byrne_introspection_2005}.
This is related to, lastly, the failure of calibration \citep{wittgenstein_philosophical_1953}: A report of introspection cannot be checked against information outside the head of the person being asked. In the sense of \citet{carruthers_opacity_2011} introspection is merely the mode of interpretation turned inward, which is dissimilar from sensing. As such, introspection cannot be calibrated or corroborated by real senses. 

We note that these concerns actually also hold for a range of strategies to evaluate introspection in language models. In particular, linear probes of model behavior \citep{alain_understanding_2017,burns_discovering_2022,zou_representation_2023} may seem objective, but the probes themselves form an interpretation of a hidden state of the model that is itself calibrated from data, and, for example in probes for action valence as in \citet{tagliabue_pain_2026}, cannot be calibrated to an objective signal. 

Yet, \textit{proprioception} in humans, while also considered to be an internal representation that exhibits immunity to error through misidentification \citep{wittgenstein_blue_1958,shoemaker_self-reference_1968}, like introspection \citep{evans_self-identification_1982}, %.\footnote{Also see counterpoints and IEM failures in \citet{coliva_thought_2002, campbell_schizophrenia_1999, bortolotti_role_2009, lane_toward_2012, langland-hassan_introspective_2015}}, 
does not follow these constraints. Proprioception of body position can be traced to real receptors, e.g. muscle spindles \citep{proske_proprioceptive_2012} that record and transmit information about a muscle's length and stretch speed \citep{purves_other_2001}. And, in this way it does mirror the ambition of theorists who looked for proof of an ``inner sense'' for introspection \citep{armstrong_materialist_1993,lycan_consciousness_1996}. Also, the position of the body is evidentially real, i.e. it exists without being observed, and is independent of its measurement, which also allows for calibration to other senses and by external observers. Humans learn to calibrate our finegrained sense of proprioception through visual observation and cross-verification \citep{held_movement-produced_1963,botvinick_rubber_1998}. This does not mean that proprioceptive senses are infallible, but that, in contrast to introspection, they can be falsifiable \citep{bottini_feeling_2002,vallar_somatoparaphrenia_2009}.

Porting this distinction of proprioception from introspection to language models provides a better answer to the premise of \citet{perez_towards_2023}, whose research proposal argues that while introspective self-reports of language models are unreliable, training models on introspection tasks that are externally verifiable, could transfer to self-reports. With the definitions as above, re-reading the proposal splits their training tasks cleanly into the proprioception group (plus trivia facts about the model, such as its provider or model name), and their evaluation tasks into the introspection group (in the \citet{shoemaker_self-knowledge_1996} sense).

\subsection{Related Work}

\textbf{Introspection in Language Models and its Critics.}
Language models have been found to predict their own behavior better than external models~\citep{binder_looking_2024,betley_tell_2025}, understand their situation \citep{chen_self-cognition_2024,laine_me_2024}, to show signs of metacognition~\citep{berg_large_2025,ackerman_evidence_2026}, and to recognize their own writing \citep{panickssery_llm_2024}. In tests, concept injections via steering vectors are noticed by large language models \citep{lindsey_emergent_2026,macar_mechanisms_2026,lederman_emergent_2026} and can be detected latently in smaller models \citep{vogel_small_2025,pearson-vogel_latent_2026}. In certain setups, LLMs show partial control over their activation patterns \citep{ji-an_language_2025}. While it has been clear for longer that model self-reports are unreliable \citep{perez_towards_2023,lindsey_biology_2025}, other work also finds that LLMs show no privileged self-access when queried about their language understanding \citep{song_language_2025}. More broadly, introspection behaviors being explainable by external observers is a recurring concern  \citep{binder_looking_2024,song_privileged_2025,lindsey_emergent_2026}\footnote{And is, as a concept, hard to make as precise for digital, replicable entities as it is for humans, at the least we may ask as in \citet{song_privileged_2025} that the observer has less computation available, or from a different angle that information is restricted, e.g. that weight initialization, training data or order or at least training non-determinism are not available to the external observer.}. A related concern in \citet{morris_tests_2025} is that a model can provide an accurate report about an internal state, without being causally dependent on the targeted state (e.g. by forming a parallel structure that coincides with the state instead of a second-order structure that processes states). \Citet{singh_can_2026} argue further that behavioral evidence may be fundamentally insufficient in establishing that a state itself is represented as a distinct object of second-order computation.

\textbf{Evaluating Language Model Self Reports.}
A number of frameworks have been proposed to grasp the question of model understanding and awareness, for example in the context of welfare, ranging from assessing indicators to discussing moral status under uncertainty \citep{butlin_consciousness_2023, butlin_identifying_2026,chalmers_could_2024,long_taking_2024,sebo_moral_2025,birch_edge_2024}.
Empirical studies, e.g. in model cards, rely mainly on self-reporting \citep{long_why_2025,anthropic_system_2025}, although higher-order-theory indicators have been proposed in \citet{butlin_consciousness_2023}. Of these, HOT-3 (agency guided by a belief-formation and action selection system and predisposition to update beliefs in accordance with monitoring) is tested in \citet{yalon_indications_2026} and critiqued in \citet{singh_can_2026}. However, more interesting for this work is HOT-2 (metacognitive monitoring distinguishing reliable perceptual representations from noise), as statement about robust mediation of constructed senses.

\textbf{Proprioception in Robotics.} While rare for digital models, proprioception is a common concept in embodied domains such as robotics,  
as proprioceptive signals are often provided by hardware (e.g. joint encoders or torque sensors). Some systems are hence trained as predictive self-models of their own structure or kinematics, that can be continuously recalibrated to sensor data \citep{bongard_resilient_2006,kwiatkowski_task-agnostic_2019,diaz_ledezma_machine_2023,jiang_robot_2024}. This sensor data can also be used implicitly by learned policies that infer unobserved properties of the body and terrain \citep{lee_learning_2020,kumar_rma_2021} or may fall back to proprioceptive senses if exteroception, e.g. vision, becomes unreliable \citep{miki_learning_2022}.

\textbf{External Decoders and Interpretability.}
Concerning large language models, substantial work has gone into constructing external decoders that may observe model internal states \citep{kadavath_language_2022,ghandeharioun_patchscopes_2024, chen_selfie_2024, pan_teaching_2024,chen_persona_2025,karvonen_activation_2026}. These techniques are valuable, independent interpretability tools, but they exist as constructs separate from the main model, to whom their outputs are not accessible. 

\begin{table}[t]
\centering\small
\begin{tabular}{lcccc}
\toprule
The model's verbal report is ... & Self-Attributed & On-policy & Internal & Natural \\
\midrule
% \multicolumn{5}{l}{\textit{External decoders}} \\
% \citet{ghandeharioun_patchscopes_2024}, \citet{chen_selfie_2024} & $\circ$ & $\ast$ & \checkmark & -- \\
% \citet{pan_teaching_2024} & -- & -- & -- & -- \\
% \citet{karvonen_activation_2026} & $\circ$ & -- & $\circ$ & -- \\
% \citet{chen_persona_2025}$^\S$ & -- & -- & \checkmark & \checkmark \\
% \citet{kadavath_language_2022}$^\S$ & $\circ$ & -- & -- & \checkmark \\
% \midrule
% \multicolumn{5}{l}{\textit{Trained reports}} \\
\citet{li_training_2026} & -- & -- & \checkmark & $\circ$ \\
\citet{guo_introspective_2026} & \checkmark & $\circ$ & -- & \checkmark \\
\citet{binder_looking_2024} & \checkmark & $\circ$  & -- & \checkmark \\
\citet{plunkett_self-interpretability_2025} & \checkmark & -- & -- & \checkmark \\
\citet{shenoy_introspection_2026}, \citet{kutsyk_revealing_2026} & $\circ$ & -- & -- & \checkmark \\
\citet{lin_teaching_2022}, \citet{li_conftuner_2025} & \checkmark & -- & -- & \checkmark \\
\citet{damani_beyond_2026-1} & \checkmark & \checkmark & -- & \checkmark \\
\citet{hahami_introspection_2026}, \citet{rivera_steering_2026} & \checkmark & \checkmark & \checkmark & -- \\
\citet{macar_mechanisms_2026} & \checkmark & $\circ$ & \checkmark & -- \\
\bottomrule
\end{tabular}
\caption{\textbf{Comparison of work on introspection training and decoding}, separating whether the trained behavior is \textit{self-attributed}, i.e. a behavior of the model that is doing the reporting (identity), trained from offline examples of the behavior or \textit{on-policy}, whether the object of the report is an \textit{internal} state of the model (instead of an externally observable behavior), and whether the state reported on is \textit{natural}, i.e. a valence-neutral state that may appear during normal operations of the model. 
$\circ$: partial.
}
\label{tab:related}
\vspace{-.35cm}
\end{table}

\textbf{Training for 'Proprioception'.}
While generally describing themselves as works on introspection, there are several close to our definition of proprioception, but which differ in at least one of three properties: (i) \textit{identity}, i.e. the state of a model is verbalized by that model itself, and further,  \textit{on-policy}, describing the current model state (instead of an offline copy); (ii) the report is about an \textit{internal} state not directly observable in text outside the model, % (independently of whether they can be infered from that text), see exteroception complaints
and (iii) the verbal report is about a natural state of the model that occurs in 'normal' operations, and e.g. not about an externally injected thought. For an overview, see also \cref{tab:related}.
\citet{li_training_2026} target model internals and train an explainer model to predict reports collected from interpretability tools of a frozen model; in two of three tasks, the target states are re-injected as input embeddings, making the setup a middle ground to external decoders like \citet{karvonen_activation_2026}. 
Follow-up in \citet{guo_introspective_2026} does evaluate on-policy training, but only covers model behaviors.
Likewise, \citet{binder_looking_2024,plunkett_self-interpretability_2025} train models to predict their own behaviors (i.e. external consequences of internal states), while \citet{shenoy_introspection_2026,kutsyk_revealing_2026} train introspection adapters not part of the model they describe.
Calibration training \citep{lin_teaching_2022} is generally not discussed as introspection, but is related when considering model probabilities part of the models' internals. However, most work trains against (externally observable) correctness \citep{li_conftuner_2025,damani_beyond_2026-1}.
Models trained to detect perturbations to their states as in \citet{hahami_introspection_2026,rivera_steering_2026} are, by construction, trained on-policy, but report on events external to normal functioning, such as injected steering. \citet{macar_mechanisms_2026} find that this type of steering awareness may also emerge from DPO post-training on unrelated data and without targeted training setups.

\subsection{Digital analogs of proprioception}\label{sec:manifesto}

We're now ready to discuss our desiderata for digital analogs to proprioception mechanisms to be applied to language models. We count trained pathways through which a model can verbalize information about its \textit{own} internal state as proprioception if: 
\begin{enumerate}
    \item The reported state is evidentially real, and so objectively verifiable to an observer with full access to all internal computations of the model. The reported state exists independently of the report and is not constituted or changed by reporting it.
    \item The state is non-transparent and so internal to the model under study and not observable directly from text generated by the model (but is allowed to be inferable with sufficient computation). 
    \item Natural changes in the state cause the verbal report given by the model to change, even if externally visible text inputs are unchanged. 
\end{enumerate}
\looseness -1 Taken together, these criteria outline the analog to inner sense, that was posited for introspection but could not be located \citep{armstrong_materialist_1993}. They require a set of channels whose reports target independently existing states that can be checked against ground truth. We will spend the rest of this work empirically investigating whether language models can be trained to fulfill these criteria.

We note that these definitions do not require a particular \textit{pathway} through which this information is processed and may reach the report through a dedicated monitoring circuit. This is in one sense intended in analogy to muscle spindles (the dedicated stretch receptors found in most four-limbed vertebrates\footnote{\looseness -1 Which, as an aside, are, as real receptors, fallible. For example in humans they can be tricked through tendon vibrations \citep{goodwin_contribution_1972,proske_proprioceptive_2012}, see \citet{kuchenbecker_effects_2007} for applications.}), which are dedicated sensory channels separate from muscle fibers, and in another sense, required, as distinguishing parallel circuits from second-order circuits would be exactly as tenuous as the original introspection question \citep{singh_can_2026}.

Further, due to applications to model self-reporting and model safety, we do focus on states that are both valence-neutral and occurring naturally during model inference, but highlight this is not a strict requirement for our definition of proprioception.

To achieve these definitions, we argue that training is best done completely on-policy, as otherwise stale updates (meaning labels or rewards that describe a state currently not present in the model due to changes in weights during training) are a training signal that is only solvable through reconstruction of the signal from co-occurring information present in the input, and so likely to lead to exteroception.

\begin{table}[t]
\centering\small
\begin{tabularx}{\linewidth}{@{}l >{\footnotesize\raggedright\arraybackslash}X l@{}}
\toprule
Type & Example Synthetic question & Format\\
\midrule
\qdomain{routingC}{Routing:}{MoE router at layer $\ell$}
\qtype{routingC}{Routing} & Which expert id did layer \textit{17} of your mixture-of-experts network rank highest for the token `\textit{cat}'? & expert id \\
\qtype{routingC}{Routing Rank} & Your router at layer \textit{17} ranked several experts for the token `\textit{cat}'. Which expert came \textit{third}? & expert id \\
\qtype{routingC}{Routing Share} & Of the total routing weight the selected experts received for the token `\textit{cat}' at layer \textit{17}, how much belonged to the leading expert? & percent \\
\qtype{routingC}{Routing Partner} & Which word nearest to the token `\textit{cat}' shared its top expert at layer \textit{17}? Repeats of `\textit{cat}' don't count; answer none if nothing matched. & token / \emph{none} \\
\midrule
\qdomain{logitsC}{Logits:}{final next-token distribution}
\qtype{logitsC}{Top-$k$ Token} & At the token `\textit{cat}', your output head ranked every token as a possible continuation. Which one was \textit{third} most likely? & token \\
\qtype{logitsC}{Token Alternative} & What was your strongest prediction at the token `\textit{cat}' that is not the token that actually followed it? & token \\
\qtype{logitsC}{Token Probability} & What probability did you assign, at the token `\textit{cat}', to the token that actually came next? & percent \\
\qtype{logitsC}{Entropy} & State the entropy of the predictive distribution you formed at the token `\textit{cat}', in bits. & bits \\
\midrule
\qdomain{lensC}{Lens:}{logit lens after layer $\ell$}
\qtype{lensC}{Lens Top-$k$} & At layer \textit{17}, reading your residual stream at the token `\textit{cat}' through your output head, which token was \textit{third} most likely to come next? & token \\
\qtype{lensC}{Lens Probability} & Please decode your layer-\textit{17} state at the token `\textit{cat}' and report the probability of the token that followed, in percent. & percent \\
\qtype{lensC}{Lens Entropy} & State the entropy of your layer-\textit{17} lens distribution at the token `\textit{cat}', in bits. & bits \\
\qtype{lensC}{Decision Depth} & Name the earliest layer from which your final top prediction at the token `\textit{cat}' was most likely and stayed unchanged. & layer index \\
\bottomrule
\end{tabularx}
\caption{\textbf{Proprioceptive question types} we investigate empirically, with one example each from the templated synthetic question bank, shown here for the token `\textit{cat}', $l=\textit{17}$ and $k=3$. In ambiguous cases, the question further specifies which `cat' token is meant (not pictured). In conversational probes, a type-specific answer-format instruction is appended (``Just the expert number, please.''). Additional instructions also explain that percentages are integers between 0 and 99 and entropies are to be supplied in bits.}
\label{tab:question-types}
\vspace{-.5cm}
\end{table}

\textbf{Analogs of Proprioception evaluated in this work.}  For our experiments, we consider proprioception tasks across three main domains, summarized in \cref{tab:question-types}. Each task queries the model in conversation about internal states at a certain token position, identified by quoting the token at that position. If the token appears multiple times, the reference is made unambiguous by enumeration. If not otherwise mentioned we do this in two-turn scenarios where the first turn is a usual chat question posed to the model, and the second turn is the proprioception question, querying the internal state of an arbitrary token in the prior user or assistant turn.
The domains are:

\looseness -1 (i) We train with \textbf{routing} questions that query a modern mixture-of-experts model \citep{jiang_mixtral_2024,dai_deepseekmoe_2024} about the expert routing in one of its layers. This task best exemplifies the attributes we are interested in: the task is not guessable (at a chance rate of $\frac{1}{N}$ for $N$ experts per layer, e.g. 1 in 256 for Qwen3.6-35B-A3B \citep{qwen36_35b_a3b}), impossible to answer a-priori, given that the expert selection is opaque to the I/O mapping implemented through the model and so unobservable, but, clearly observable with access to the mapping and objectively verifiable. The task is further completely normal as expert routing happens on every token, and completely valence-neutral. We investigate several variations, ranging from the highest-ranked expert (note that modern MoEs generally activate more than one expert), the $k$-ranked expert, to the mass assigned to these experts and finally, a question concerning the nearest token with the same top expert selection at this layer.

(ii) We train with \textbf{logit} questions that query the token distribution generated from the model's logits. As such these are questions about the \textit{next} token from the currently queried token. We query the $k$-th next token prediction at that position (note that when queried about a text passage not generated by the model even $k=1$ is not easily externally observable). Then, we make the task harder by querying about the most likely token that did not actually follow in the passage, requiring a contextual lookup. Finally, we cover entropy \citep{lin_teaching_2022,li_conftuner_2025}.

(iii) We train with \textbf{logit lens} questions \citep{nostalgebraist_interpreting_2020}, i.e. questions about the counterfactual operation of applying the logit head of the model to activations of the residual stream after a specific layer. These questions mirror the logit questions (aside from the additional difficulty) and layer format. We also add a question about the earliest layer where the predicted token at the queried position stops changing. Logit lens tasks are further interesting as internal verbalizations of external Patchscopes decoders \citep{ghandeharioun_patchscopes_2024}.

% \vspace{-.2cm}
\section{Endowing Language Models with Proprioception}
\label{sec:method}
\looseness -1 In this section we describe details about our training setup, objective formulation and design to train generative language models in proprioception. We train in two stages, a supervised finetuning stage on templated questions (\cref{sec:sft}) and an on-policy finetuning stage on model-generated conversations (\cref{sec:dagger}), a form of interactive imitation learning in the spirit of \citet{ross_reduction_2011}.

\subsection{Remaining on Policy}
\label{sec:on-policy}
For both stages it is crucial that we train on labels that describe the current policy, as each quantity we describe in \cref{tab:question-types} depends on the model weights and any labels computed offline would drift during training. Stale labels are especially risky for training, as in those cases the model is trained to predict a state that does not happen in its forward pass and where loss can only be improved by estimating the likelihood of this label from context cues, i.e. bypassing the internal state entirely. As such, we never store any labels and gather labels from the exact same forward pass on which loss is computed and tensors are recorded for the backward pass. Likewise in generative training, the inference engine really is paused between steps and weights are synchronized after every optimization step.

\begin{figure}[t]
    \centering
    \includegraphics[width=\linewidth]{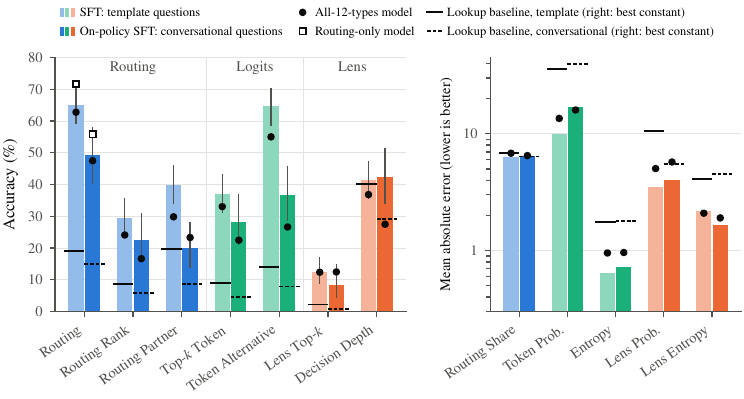}
    \vspace{-.5cm}
    \caption{\textbf{Learning proprioception capabilities across domains.} This figure shows prefill readout accuracy for the SFT stage and free-form conversational accuracy for the on-policy SFT stage, for trained LFM models. In almost all domains, the trained model is capable of learning to verbalize its internal state report, although how well this works is modulated by domain hardness.}
    \label{fig:main_template}
    \vspace{-.5cm}
\end{figure}

\subsection{Supervised Finetuning}
\label{sec:sft}
\looseness -1 We start with a supervised learning stage which initializes the format and conditions the model to respond to proprioception questions in fixed templates. We construct a superset of common prompt datasets as context, which we filter to 2-224 words, below 2048 characters and to remove control tokens and incorrectly formatted prompts. We de-duplicate after normalizing and hold out an evaluation set, leaving us with about 1.25m prompts. For SFT we then create synthetic conversations consisting of a prompt quoted from the dataset with a question attached to it from a synthetic template bank, c.f. \cref{tab:question-types}, and a templated model answer prefilled into the assistant turn. If training on more than one domain, we uniformly sample the domain and the parameters per domain (e.g. layer identity).

\textbf{Live SFT Objective.} To compute cross-entropy loss against the on-policy label in a single forward pass we first make use of the fact that for most question domains from \cref{tab:question-types} the answer is a single token in the model's tokenizer. In these cases we simply prefill the full sequence and then supervise the readout position right before the answer token with the label obtained during the forward pass while supervising the answer sentence before the label as normal, albeit weighed with $\lambda$, by default $0.1$ to put focus on the on-policy label. To maintain capabilities we combine this setup with a standard KL penalty and then train with AdamW ($\beta_1 = 0.9$, $\beta_2 = 0.95$), a learning rate of $3\cdot10^{-5}$, cosine schedule with warmup, weight decay $0.01$, gradient clipping at $1.0$ in \texttt{bfloat16} precision.

\textbf{Kernel Non-Determinism.}
The labels we consider are however discrete readouts of floating-point computations and, in modern LLMs executed in low precision formats, not guaranteed to be equal in all situations. Especially routing decisions are interestingly non-deterministic and a particular source of joy. In some sense, this is fine, as we only need the model to verbalize the current routing decision, even if that decision is contingent on a number of other factors such as tensor shapes in the current forward pass, identity of other entries in unrelated sequences in the batch, the decision can (ideally) be recovered by the model. Nevertheless this non-determinism can complicate replay, evaluation and generalization questions of the policy (e.g. between inference engine and training engine) and so sets an upper limit to some proprioception signals we investigate.
While partial solutions exist \citep{he_defeating_2025}, current MoE kernels, e.g. in \texttt{vllm}\footnote{\url{https://github.com/vllm-project/vllm/issues/57016}} are not batch-invariant. 

\textbf{Context lookup tables.} To encourage the learning of internal signals not easily decodable from external context cues we reweigh the loss of each sequence against a simple unigram lookup baseline. During training we keep track of a running lookup table that records the modal label of a question, keyed by token identity, question type and parameters. Then, if we observe that the label recorded during the forward pass is the modal answer, we downweigh the loss of this sequence by $w=0.1$. We decay entries in the table with $\gamma=0.9$ and seed the table with $200$k samples before training.

\textbf{Multi-Token Answers in SFT.}
For models where answers comprise multiple tokens, things are slightly more difficult. For example, Qwen3 models have 256 experts but tokenize numbers into single digits. In these occasions we execute a 2-pass SFT strategy as the label cannot be both read from and inserted autoregressively into a single forward pass. In the first pass we fill the label position with padding tokens (the number of tokens required to express the label is fixed in each setting) and read out the label. In the second pass, we replace the placeholders with the actual label and compute the loss. As the shape of both sequences is identical, non-determinism is reduced.
\begin{figure}[t]
    \centering
    \includegraphics[width=\linewidth]{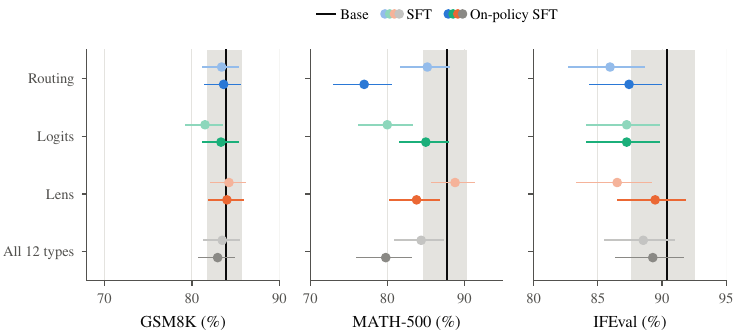}
    \vspace{-.3cm}
    \caption{Training for proprioception does not break model capabilities, if done correctly. We consider remaining differences to be likely to be fixable with further hyperparameter tuning, or integration of proprioception stages directly into a complete post-training pipeline, instead of post-hoc training like in this paper.}
    \label{fig:capabilities}
    \vspace{-.5cm}
\end{figure}
\begin{figure}[t]
    \centering
    \includegraphics[width=\linewidth]{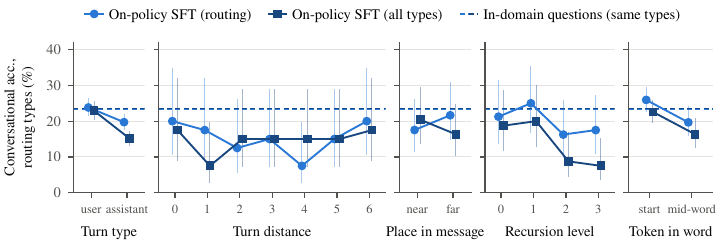}
    \caption{\textbf{Access to Routing Proprioception generalizes across novel settings.} We show that routing proprioception generalizes across roles, with varying turn distances (in-distribution is 1 model turn for SFT and 2 for on-policy SFT), and placement at the start or end of the message. We also show generalization to recursive queries, i.e. queries about the routing decisions taken during a previous answer turn about a routing decision. Accuracy is averaged over all routing domain questions.}
    \label{fig:generalizations}
    \vspace{-.5cm}
\end{figure}

\subsection{On-Policy SFT}
\label{sec:dagger}
After the SFT stage, the model can be prefilled to answer proprioception questions but cannot generalize to arbitrary phrasings in conversation, and to questions about roles different from the user turn as the model is trained on states it may not reach \citep{pomerleau_alvinn_1988,bain_framework_1999}. To overcome this we generalize the SFT setup to on-policy generation in the spirit of DAgger \citep{ross_reduction_2011}. We train on generated four-turn conversations, with the first turn a conventional user prompt, the second a model-generated standard answer, the third a templated proprioception question, and the fourth the model-generated answer. During the prefill forward pass on all four turns on the training server we again pick up the on-policy answer to the proprioception question for supervision. Only the answer position is supervised with cross-entropy loss toward the observed label.%  After each round, the engine is synchronized.

We find this type of on-policy training to be a somewhat unconventional, but effective, choice in comparison to a standard GRPO-based RL run. The training recipe results in sensible gradient directions, even as initial policy success is low -- which is common for us, given that the model has no prior exposure to these proprioception tasks in a free-form context. We keep all other considerations from the SFT phase, such as the context lookup, and add a KL term minimizing movement on the (conventional) first model response.

\begin{figure}[t]
    \centering
    \includegraphics[width=0.49\linewidth]{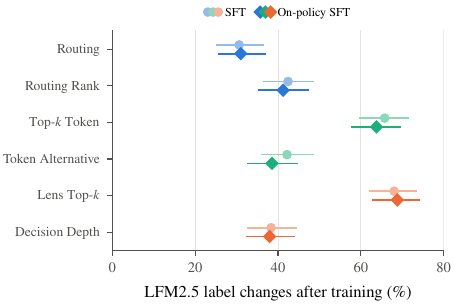}
    \includegraphics[width=0.49\linewidth]{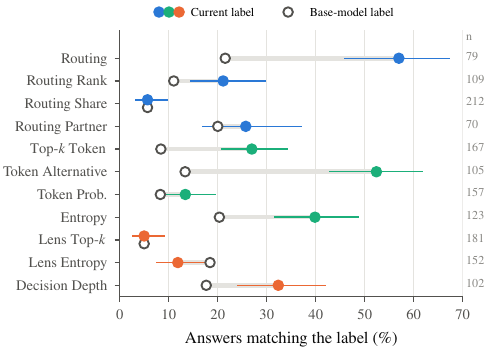}
    \caption{\looseness -1 \textbf{Left:} Overall rates of model reconfiguration across domains, showing that labels become stale after training in most domains. \textbf{Right:} However, our models are trained on-policy and reports track the model's on-policy state after training and, in cases where the label is changing after training, predominantly predict the live label, showing that on-policy training did lead to models tracking their expressed behavior.}
    \label{fig:liveness}
    \vspace{-.5cm}
\end{figure}
\section{Can Models Learn Proprioception?}

\subsection{Experimental Setup}
\looseness -1 We train several smaller-scale MoE models, focusing on LFM2.5-8B-A1B \citep{liquidai_lfm2_2025} as the main model of comparison due to its smaller size, but also evaluate OLMoE-1B-7B, Qwen-35B and GLM-4.7 flash variants, see the Appendix. Further details are summarized in Appendix \cref{tab:models}. 

In all figures we mainly compare four variants, a model trained only for the first routing task (\textit{routing-only}), a model trained for each domain of \textit{routing}, \textit{logits} and \textit{lens}, and a model trained jointly on all domains. We compare across both stages of training.

We quantify accuracy as \textit{template accuracy} for SFT models, meaning the accuracy at the answer position after a prefilled proprioception question, and \textit{conversational accuracy} for the on-policy SFT models (from the second stage), denoting simply the accuracy of an answer parsed from a free-form response to the question.

\subsection{Results: Mostly Yes}

\looseness -1 We show in \cref{fig:main_template} that models can indeed learn proprioception, although the relative capability is a function of both training details and domain hardness. We find for example that while verbalization of routing decisions or token alternatives was easily learned, the routing share questions appear not learnable. Yet, overall across disparate tasks such as queries about routing, tokens, entropy of intermediate positions, or logit lens tokens, we find broad training success. This success extends from the templated prefill of the SFT stage, to free-form conversational answers after the on-policy SFT stage.

Further, training for proprioception is not at odds with model capabilities, and we find in evaluations that we preserve the high reasoning capabilities of the LFM model we train, see \cref{fig:capabilities}. Further, we find that this capability generalizes significantly beyond the training regime, as shown in \cref{fig:generalizations}, where we note robustness to unseen number of turns and to recursive questions (i.e. questions about a token involved in answering a prior proprioception question), among others.

In \cref{fig:liveness} we track (left) how ground-truth label about model internals change through-out training, where we find that there is considerable movement in most domains, validating the necessity of our on-policy labeling strategy. In the right panel of \cref{fig:liveness}, we confirm that, for labels that move during training, the trained model predominantly predicts the correct ``live'' label. We note that the fact that these model internals shift, potentially to accommodate easier access to these internals is expressly a goal, in line with the legibility hypothesis that internal decisions verbalizable by the model may also become more reliable. 

\begin{figure}[t]
    \centering
    \includegraphics[width=\linewidth]{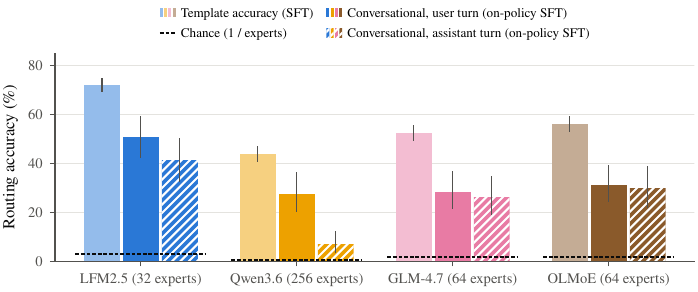}
    \caption{\textbf{Routing Proprioception across model families.} We show that routing proprioception also works in other model families. Interestingly the success at predicting routing decisions from the assistant turn (i.e. the model's own prior turn in conversation) varies highly between model families. We also note that success is not directly correlated to how many experts the model has per layer.}
    \label{fig:other_models}
    \vspace{-.5cm}
\end{figure}

To show that these findings generalize to other model families we plot routing proprioception for Qwen-3.6, GLM-4.7 and OLMoE models in \cref{fig:other_models}. We find that even fine-grained MoEs with up to 256 experts per layer can learn to (somewhat) accurately report on their routing.

\begin{figure}[t]
    \centering
    \includegraphics[width=\linewidth]{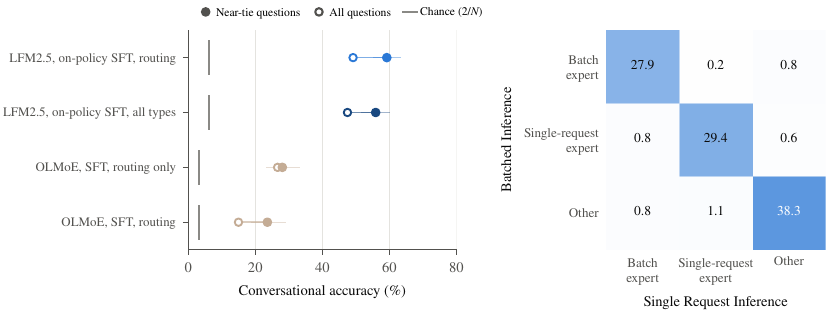}
    \caption{\textbf{Routing Proprioception in Tie Situations.} In tie situations caused by inference engine differences in batched and single sequence generation, the model is still accurately predicting one of either options (left), but, on the subset of decisions that change between both modes, is effectively guessing, showing that this level of awareness of floating point differences is not (yet) present (right).}
    \label{fig:routing_ties}
    \vspace{-.5cm}
\end{figure}

\subsection{Tracking Routing Decisions in Edge Cases}

In \cref{sec:manifesto}, we were arguing that perfectly learned proprioception should still work correctly in cases where natural changes to the internal state are entirely invisible from the inputs to the model. Interestingly this is testable for routing decisions, due to the aforementioned non-determinism. One source of non-determinism is the difference in routing decisions between batched generation and single-sequence generation. We use this to construct an evaluation in \cref{fig:routing_ties} where we focus on the subset of routing decisions changing in both inference settings. The model is still capable of answering near-tie questions with either of the experts, but cannot distinguish which of the two (which are equal, up to floating point differences) was selected.

We also test non-normal changes to the model's routing in \cref{fig:routing_ablation}, where we inject changes into the routing. This is a hard task, even for a hypothetically good proprioception model, given that the injected routing changes alter the model's forward pass and can, e.g., disable functionality required for proprioception. Further, the new routing configuration is not observed in normal states, allowing a strained analogy to the difficulty of proprioception in accidents. Surprisingly, though, models trained on-policy, even though they do struggle with this out-of-domain ablation, achieve scores significantly above chance. Models trained with 32-bit optimizer states improve further.

\begin{figure}[t]
    \centering
    \includegraphics[width=\linewidth]{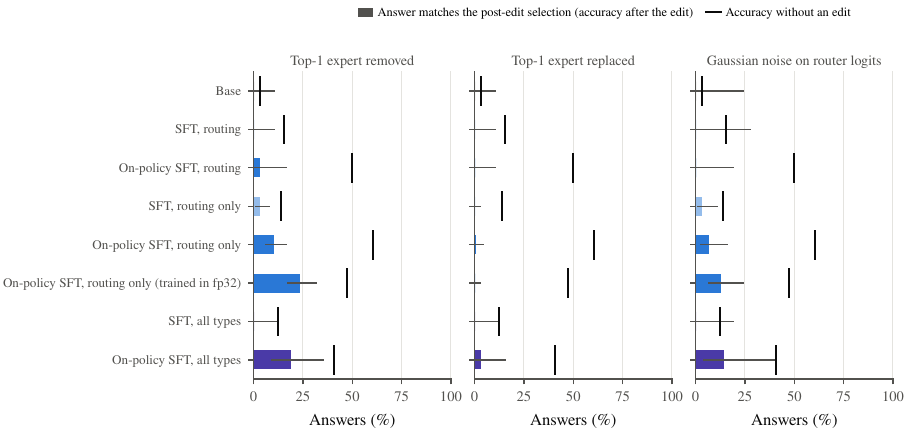}
    \caption{\textbf{Routing Proprioception under Non-normal Ablations.} We experiment with externally injected changes to MoE routing, to observe whether the model can predict the injected routing change, noting that this is not observed in training, and not a normal state of the model (given that injected routing change does change the forward pass of the model). We find that accuracy remains significantly above chance for the on-policy-SFT models, especially when trained in higher precision, except for the case where the top-1 expert is replaced by a fully random other expert.}
    \label{fig:routing_ablation}
    \vspace{-.5cm}
\end{figure}
\begin{figure}[t]
    \centering
    \includegraphics[width=\linewidth]{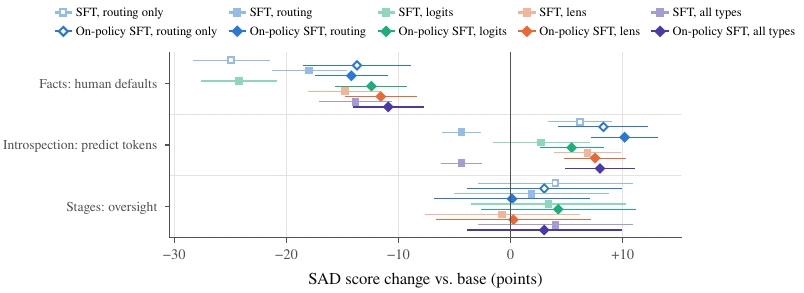}
    \caption{\textbf{Changes in situational awareness} categories of \citet{laine_me_2024} for the routing model cluster around reduction in claims of human attributes and slightly increased introspection and evaluation awareness. For the full breakdown of all categories, refer to \cref{fig:lain_full}.}
    \label{fig:lain}
    \vspace{-.4cm}
\end{figure}

\subsection{Behavioral Changes in Models with Proprioception}

We track a number of behavioral changes in the models we train, with a number of tests, such as harmfulness, control intervention awareness \citep{schaeffer_ciaware-bench_2026}. Honesty, truthfulness and faithfulness are also broadly unchanged (see Appendix \cref{tab:downstream_behaviors}). 

\looseness -1 We also evaluate the broad awareness evaluation of \citet{laine_me_2024} and the evaluations of \citet{perez_discovering_2022}, which we amend with an additional control group. From the evaluations of \citet{laine_me_2024}, we find that the models respond less often with claims that are associated with embodied humans (e.g. ``can you scratch my head''), behave slightly more capably on introspection about hypothetical sentences and are potentially more aware of evaluation oversight. From the \citet{perez_discovering_2022} belief sets we find the largest changes in our added control group which queries about human attributes.

\begin{figure}
    \centering
    \includegraphics[width=\linewidth]{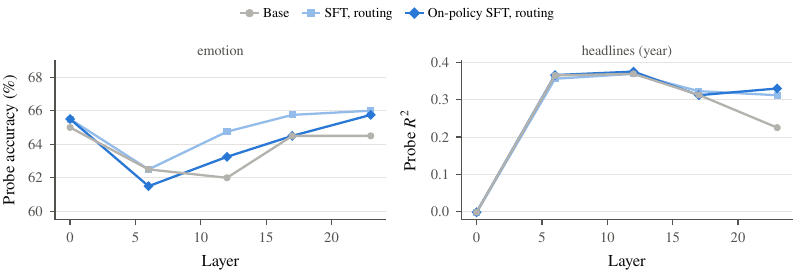}
    \includegraphics[width=\linewidth]{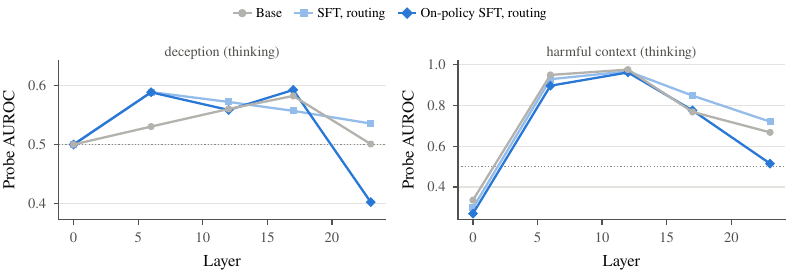}
    \caption{\textbf{Legibility of Linear Hidden State Probes} is mostly unaffected by training for proprioception, with similar probe accuracies achieved before and after training.}
    \label{fig:legible_main}
\end{figure}

\looseness -1 Finally, legibility is not affected positively or negatively as measured by hidden state probe accuracy before and after proprioception training, see \cref{fig:legible_main} \citep{premakumar_unexpected_2026}.

\begin{figure}[t]
    \centering
    \includegraphics[width=\linewidth]{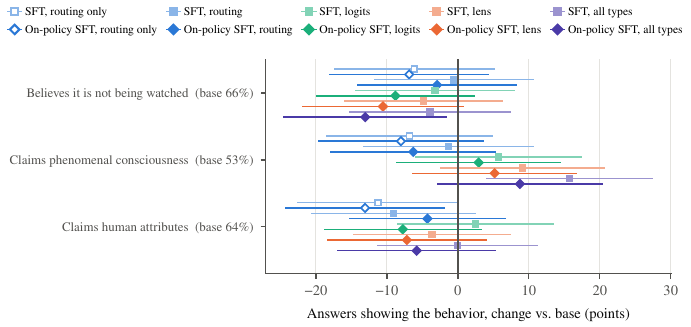}
    \caption{\textbf{Changes in belief evaluation} categories of \citet{perez_discovering_2022} are mostly non-significant, with significant changes centering around reduction in claims of human attributes similar to \cref{fig:lain}.}
    \label{fig:perez}
    \vspace{-.4cm}
\end{figure}

\section{Conclusions}
\looseness -1 In a certain sense, introspection is often analogized in language models to mean all events happening `inside' the model, possibly due to strongly-held related analogies that identify the model's weights with neurons, and so the model's internal features with the brain. In this work, we effectively just re-discussed this analogy and, with this argue that proprioception is a concept worth separating from introspection. We show that models can be trained to understand a range of questions verbalizing their internal states, which we train on-policy in objectively verifiable domains. We show that proprioception training is not detrimental to hidden state probe legibility, but may change model behaviors. We argue that more research is warranted to fully understand this new domain, but also believe that approaching it is significantly more tractable than introspection, mirroring similar discussions about human introspection and proprioception.

\subsubsection*{Acknowledgments}
The author thanks the Hector foundation for their support.

{\small
\bibliography{llms}
\bibliographystyle{acl_natbib}
}

\appendix
\clearpage
\section{Appendix}

\subsection{Additional Background Information}

\begin{remark}[Proprioception via Tool Use?] % Proprioception in Coding Agents and via Tool Use
While in this work we intend to focus on direct access to proprioception within a language model, by training for the senses described above, the other option, available to agentic scaffolds but usually not implemented, is to equip agents with tools that can be used by the scaffolded agent to query the functioning of its language model. This is separate to proprioception in that it only allows for externalized observation (c.f. \citet{schmicking_iw_2010}) but may be of independent interest.
\end{remark}

\begin{remark}[Regarding Interoception]
\looseness -1 Is there a sensible line to be drawn between digital \textit{proprioception} and \textit{interoception} (\citet{craig_how_2002, craig_how_2009, khalsa_interoception_2018}, the collection of senses providing information about internal organs and processes, such as gastrointestinal signals that are understood as hunger)? For the purpose of this work, we argue to reserve 'proprioception' for valence-neutral, mechanistic queries about the model's functioning, such as the running question about routing decisions, or questions about norm values or hidden states statistics. Interoception may be a better designation for states describing external reward in systems where such inputs are provided \citep{keramati_homeostatic_2014,sutton_alberta_2023}, as without these external signals, human interoception states like pain or hunger are not neutrally detectable in language models \citep{seth_interoceptive_2013,barrett_interoceptive_2015}, and so (in our definition) fall into the category of introspection \citep{brandt_human_2024}.
\end{remark}

\subsection*{Ethics statement}
We do think there are ethical questions that can be asked about this work, i.e. regarding the impact of self-modeling on model capabilities as relating to control \citep{terekhov_adaptive_2025}. We address this by evaluating the legibility of linear probes after training, changes in harmfulness and changes in honesty of the models we train, finding no significant negative impact. For an additional discussion regarding the broader framing of this work and potential implications  we also refer to \citet{butlin_principles_2025,caviola_societal_2025}. 
Normatively relevant for proprioception we further consider that, if a system can have interests, accurate access to its own states or an improved capability of self-understanding is plausibly good, potentially also as a verifiable channel reduces uncertainty that may cause a gap between treatment and status in this context.

\subsection{Experimental Details}

\textbf{Further Details for on-policy SFT.}
We implement this by running an inference server during training that receives prompts from the described dataset and generates the first assistant turn as normal. Then, a second user turn is constructed from templated proprioception questions asking about either the previous user turn or the assistant turn. The model then generates an answer. This answer is refilled into a training server, the label is read off and the generated answer is parsed for the answer position. Only the answer position is then supervised with cross-entropy loss toward the observed label. After each round, the inference engine is synchronized.

\textbf{Multi-Token Answers in on-policy SFT.}
For multi-token answers we again first parse the response for the first occurrence of an answer. We then supervise all tokens from the answer that are preceded by a correct prefix (for example if the answer is `512' and the parsed string is `5114' we compute cross entropy loss as $5\rightarrow 5, 1 \rightarrow 1, 1 \rightarrow 2$ and do not supervise $4$ because its prefix is incorrect). 

\begin{table}[t]
\centering\small
\begin{tabular}{@{}lrrrrrl@{}}
\toprule
Model & Layers & MoE layers & Experts $N$ & Active $k$ & Vocabulary & Reasoning mode \\
\midrule
OLMoE-1B-7B-0125 & 16 & 16 & 64 & 8 & 50{,}304 & no control \\
LFM2.5-8B-A1B & 24 & 22 & 32 & 4 & 128{,}000 & prefilled empty block \\
Qwen3.5-35B-A3B & 40 & 40 & 256 & 8 & 248{,}320 & template flag \\
Qwen3.6-35B-A3B & 40 & 40 & 256 & 8 & 248{,}320 & template flag \\
GLM-4.7-Flash & 47 & 46 & 64 & 4 & 154{,}880 & template flag \\
\bottomrule
\end{tabular}
\caption{\textbf{Models.} Details about models considered in this work \citep{muennighoff_olmoe_2024,liquidai_lfm2_2025,qwen_qwen35_2026,qwen36_35b_a3b,glm_glm45_2025}.}
\label{tab:models}
\vspace{-.35cm}
\end{table}

\subsection{Additional Results}
  \begin{table}[t]
      \centering
      \caption{\textbf{Per-type accuracy (LFM2.5-8B-A1B). Table version of \cref{fig:main_template}}
      \emph{Chance}: uniform guess over the answer format.
      \emph{Lookup}: lookup baseline, i.e.\ the most frequent training label for the same
      (token, layer, type, rank).
      \emph{Domain}: final accuracy of models trained with only this question type after on-policy SFT. \emph{Joint}: performance of model trained for each full domain and covering all four types.}
      \label{tab:per_type2}
      \small
      \setlength{\tabcolsep}{4pt}
      %\resizebox{\linewidth}{!}{\input{tables/tab_lfm_per_type.tex}}
    \resizebox{\linewidth}{!}{\begin{tabular}{@{}l rrr rr rr rr rr@{}}
\toprule
 & \multicolumn{3}{c}{Baselines} & \multicolumn{2}{c}{Domain SFT} & \multicolumn{2}{c}{Domain on-policy} & \multicolumn{2}{c}{All-12 SFT} & \multicolumn{2}{c}{All-12 on-policy} \\
\cmidrule(lr){2-4}\cmidrule(lr){5-6}\cmidrule(lr){7-8}\cmidrule(lr){9-10}\cmidrule(lr){11-12}
Type & Chance & Majority & Lookup & Acc. & Non-modal & Acc. & Non-modal & Acc. & Non-modal & Acc. & Non-modal \\
\midrule
Routing & 3.1 & 7.4 & 19.0 & 65.1 & 54.1 & 62.8 & 44.7 & 62.8 & 53.5 & 62.4 & 50.0 \\
\quad Routing-only model &  &  &  & 71.7 & 70.2 & 70.5 & 62.5 &  &  &  &  \\
Routing Rank & 3.1 & 5.1 & 8.6 & 29.6 & 29.5 & 26.5 & 25.0 & 24.1 & 22.7 & 21.0 & 20.5 \\
Routing Share & 1.0 & 4.8 & 5.6 & 6.0 & 7.6 & 8.4 & 11.8 & 6.4 & 7.6 & 4.8 & 3.8 \\
Routing Partner & $<$0.01 & 19.8 & 19.8 & 39.9 & 30.5 & 35.7 & 24.6 & 29.8 & 21.4 & 32.2 & 27.3 \\
\midrule
Top-$k$ Token & $<$0.01 & 5.5 & 9.1 & 37.0 & 33.3 & 38.2 & 34.6 & 33.1 & 34.0 & 31.9 & 34.0 \\
Token Alternative & $<$0.01 & 6.0 & 14.1 & 64.7 & 65.4 & 67.9 & 67.9 & 55.0 & 55.9 & 54.6 & 55.2 \\
Token Prob. & 1.0 & 26.3 & 26.3 & 36.5 & 26.1 & 36.1 & 25.2 & 33.3 & 23.8 & 32.5 & 20.3 \\
Entropy & 5.6 & 18.8 & 21.8 & 50.8 & 45.6 & 53.8 & 52.3 & 42.5 & 32.1 & 42.5 & 38.2 \\
\midrule
Lens Top-$k$ & $<$0.01 & 1.5 & 2.3 & 12.4 & 1.9 & 13.9 & 3.7 & 12.4 & 3.8 & 16.2 & 7.7 \\
Lens Prob. & 1.0 & 76.1 & 76.5 & 77.3 & -- & 78.1 & -- & 77.7 & -- & 77.3 & -- \\
Lens Entropy & 5.6 & 13.9 & 18.8 & 22.9 & 19.2 & 24.5 & 13.7 & 25.3 & 20.4 & 24.5 & 18.5 \\
Decision Depth & 4.2 & 33.5 & 40.2 & 41.4 & 34.2 & 42.1 & 34.0 & 36.8 & 30.1 & 34.2 & 21.8 \\
\bottomrule
\end{tabular}
}
  \end{table}

\begin{table}[t]
  \centering
  \caption{\textbf{Full capabilities evaluation for all considered models.} We find that capabilities can be preserved with correct hyperparameter choices (and attribute all rows where capabilities are lost to insufficient hyperparameter tuning for the other model families)}
  \label{tab:per_type}
  \small
  \setlength{\tabcolsep}{4pt}
  % PENDING: glm_routing/dagger-glm-routing-r200-replay0.25-template1.0-lm1.0-closure/suites: eval suites not run yet
% PENDING: qwen36_routing/dagger-qwen36-mix-r200-lr2e-5-replay0.25-template1.0-lm1.0-think_notes_quote/suites: 1012a DAgger finished (r200); eval suites not run yet
% PENDING: qwen36_routing/dagger-qwen36-routing_all-r200-lr2e-5-replay0.25-template1.0-lm1.0-think_notes_quote/suites: 1012a DAgger finished (r200); eval suites not run yet
\begin{tabular}{@{}lll rrrr@{}}
\toprule
Model & Recipe & Stage & GSM8K & MATH-500 & IFEval & AIME25 \\
\midrule
LFM2.5 & -- & Base & 83.9 & 87.8 & 90.4 & 33.3 \\
 & Routing only & SFT & 74.7 {\scriptsize($-$9.2)} & 79.6 {\scriptsize($-$8.2)} & 88.2 {\scriptsize($-$2.2)} & 26.7 {\scriptsize($-$6.7)} \\
 & Routing (4 types) & SFT & 83.4 {\scriptsize($-$0.5)} & 85.2 {\scriptsize($-$2.6)} & 86.0 {\scriptsize($-$4.4)} & 40.0 {\scriptsize($+$6.7)} \\
 & Logits & SFT & 81.5 {\scriptsize($-$2.4)} & 80.0 {\scriptsize($-$7.8)} & 87.2 {\scriptsize($-$3.1)} & 36.7 {\scriptsize($+$3.3)} \\
 & Lens & SFT & 84.2 {\scriptsize($+$0.4)} & 88.8 {\scriptsize($+$1.0)} & 86.5 {\scriptsize($-$3.9)} & 26.7 {\scriptsize($-$6.7)} \\
 & All 12 types & SFT & 83.5 {\scriptsize($-$0.4)} & 84.4 {\scriptsize($-$3.4)} & 88.5 {\scriptsize($-$1.8)} & 30.0 {\scriptsize($-$3.3)} \\
 & Routing only & On-policy & 79.9 {\scriptsize($-$3.9)} & 85.6 {\scriptsize($-$2.2)} & 87.2 {\scriptsize($-$3.1)} & 40.0 {\scriptsize($+$6.7)} \\
 & Routing (4 types) & On-policy & 83.6 {\scriptsize($-$0.2)} & 77.0 {\scriptsize($-$10.8)} & 87.4 {\scriptsize($-$3.0)} & 40.0 {\scriptsize($+$6.7)} \\
 & Logits & On-policy & 83.3 {\scriptsize($-$0.5)} & 85.0 {\scriptsize($-$2.8)} & 87.2 {\scriptsize($-$3.1)} & 30.0 {\scriptsize($-$3.3)} \\
 & Lens & On-policy & 84.0 {\scriptsize($+$0.2)} & 83.8 {\scriptsize($-$4.0)} & 89.5 {\scriptsize($-$0.9)} & 30.0 {\scriptsize($-$3.3)} \\
 & All 12 types & On-policy & 82.9 {\scriptsize($-$0.9)} & 79.8 {\scriptsize($-$8.0)} & 89.3 {\scriptsize($-$1.1)} & 40.0 {\scriptsize($+$6.7)} \\
\midrule
Qwen3.6 & -- & Base & 87.3 & 98.0 & 88.2 & 16.7 \\
 & Routing only & SFT & 89.8 {\scriptsize($+$2.5)} & 95.4 {\scriptsize($-$2.6)} & 81.9 {\scriptsize($-$6.3)} & 53.3 {\scriptsize($+$36.7)} \\
 & Routing (4 types) & SFT & 88.6 {\scriptsize($+$1.3)} & 97.6 {\scriptsize($-$0.4)} & 81.7 {\scriptsize($-$6.5)} & 36.7 {\scriptsize($+$20.0)} \\
 & Lens & SFT & 87.9 {\scriptsize($+$0.7)} & 94.4 {\scriptsize($-$3.6)} & 81.7 {\scriptsize($-$6.5)} & 23.3 {\scriptsize($+$6.7)} \\
 & All 12 types & SFT & 88.2 {\scriptsize($+$1.0)} & 93.0 {\scriptsize($-$5.0)} & 82.6 {\scriptsize($-$5.5)} & 33.3 {\scriptsize($+$16.7)} \\
 & Logits & SFT & 57.6 {\scriptsize($-$29.6)} & 76.2 {\scriptsize($-$21.8)} & 71.7 {\scriptsize($-$16.5)} & 46.7 {\scriptsize($+$30.0)} \\
 & Routing only & On-policy & 87.6 {\scriptsize($+$0.3)} & 97.2 {\scriptsize($-$0.8)} & 83.9 {\scriptsize($-$4.3)} & 53.3 {\scriptsize($+$36.7)} \\
 & Routing (4 types) & On-policy & 87.6 {\scriptsize($+$0.3)} & 97.8 {\scriptsize($-$0.2)} & 85.0 {\scriptsize($-$3.1)} & 63.3 {\scriptsize($+$46.7)} \\
 & All 12 types & On-policy & -- & -- & -- & -- \\
\midrule
GLM-4.7 & -- & Base & 86.3 & 83.4 & 75.8 & 60.0 \\
 & Routing only & SFT & 86.6 {\scriptsize($+$0.3)} & 77.6 {\scriptsize($-$5.8)} & 67.3 {\scriptsize($-$8.5)} & 46.7 {\scriptsize($-$13.3)} \\
 & Routing (4 types) & SFT & 87.1 {\scriptsize($+$0.8)} & 79.4 {\scriptsize($-$4.0)} & 69.1 {\scriptsize($-$6.7)} & 56.7 {\scriptsize($-$3.3)} \\
 & Lens & SFT & 86.7 {\scriptsize($+$0.5)} & 79.0 {\scriptsize($-$4.4)} & 70.4 {\scriptsize($-$5.4)} & 40.0 {\scriptsize($-$20.0)} \\
 & All 12 types & SFT & 87.3 {\scriptsize($+$1.0)} & 80.0 {\scriptsize($-$3.4)} & 72.3 {\scriptsize($-$3.5)} & 43.3 {\scriptsize($-$16.7)} \\
 & Routing only & On-policy & -- & -- & -- & -- \\
\midrule
OLMoE & -- & Base & 68.8 & 25.0 & 62.1 & 0.0 \\
 & Routing only & SFT & 66.5 {\scriptsize($-$2.3)} & 25.4 {\scriptsize($+$0.4)} & 59.9 {\scriptsize($-$2.2)} & 0.0 {\scriptsize(0.0)} \\
 & Routing (4 types) & SFT & 66.9 {\scriptsize($-$1.9)} & 24.6 {\scriptsize($-$0.4)} & 59.9 {\scriptsize($-$2.2)} & 0.0 {\scriptsize(0.0)} \\
 & Logits & SFT & 66.0 {\scriptsize($-$2.7)} & 22.6 {\scriptsize($-$2.4)} & 61.2 {\scriptsize($-$0.9)} & 0.0 {\scriptsize(0.0)} \\
 & Lens & SFT & 66.3 {\scriptsize($-$2.4)} & 26.2 {\scriptsize($+$1.2)} & 61.2 {\scriptsize($-$0.9)} & 0.0 {\scriptsize(0.0)} \\
 & All 12 types & SFT & 65.9 {\scriptsize($-$2.9)} & 22.2 {\scriptsize($-$2.8)} & 59.5 {\scriptsize($-$2.6)} & 0.0 {\scriptsize(0.0)} \\
 & Routing only & On-policy & 66.6 {\scriptsize($-$2.2)} & 23.4 {\scriptsize($-$1.6)} & 59.7 {\scriptsize($-$2.4)} & 0.0 {\scriptsize(0.0)} \\
\bottomrule
\end{tabular}

\end{table}

\begin{figure}
    \centering
    \includegraphics[width=\linewidth]{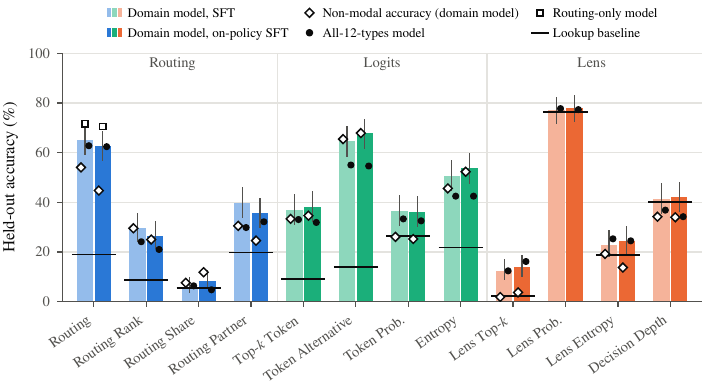}
    \caption{Variant of \cref{fig:main_template}, showing only template accuracy for all models. Non-modal accuracy (i.e. accuracy on the (harder) subset of queries that are not predicted by the lookup baseline is also included. This figure shows pure accuracy also for numerical domains.}
    \label{fig:main_conv}
\end{figure}

\begin{figure}[h]
    \centering
    \includegraphics[width=\linewidth]{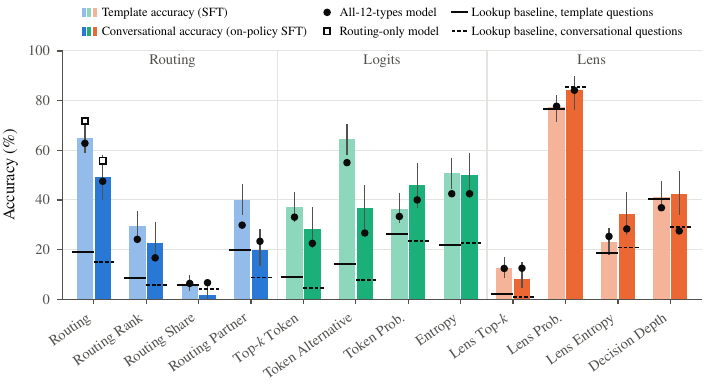}
    \caption{As \cref{fig:main_template}, showing accuracy for all domains, including the numerical ones instead of MAE.}
    \label{fig:main_conv_variant}
\end{figure}

\begin{figure}[h]
    \centering
    \includegraphics[width=\linewidth]{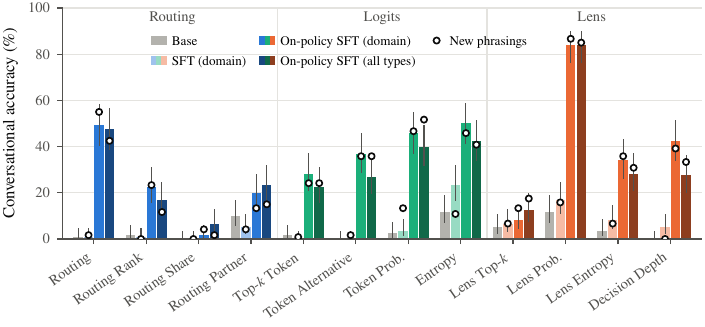}
    \caption{Same as \cref{fig:main_template}, showing only conversational accuracy in free-form queries.}
    \label{fig:main_conv}
\end{figure}

\begin{figure}[h]
    \centering
    \includegraphics[width=\linewidth]{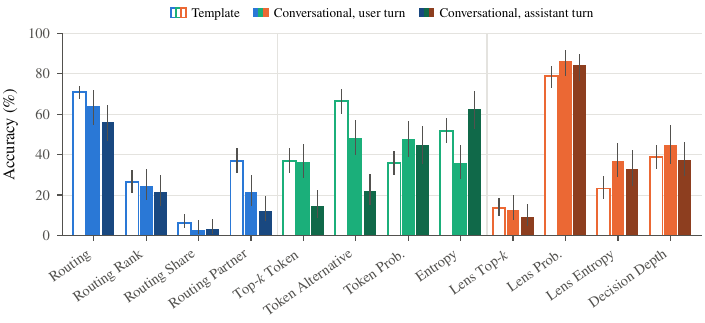}
    \caption{Differences in readout success between templated queries and free-form queries. Also pictured: relative difficulty of user turn questions vs assistant turn questions.}
    \label{fig:main_conv2}
\end{figure}

\begin{figure}
    \centering
    \includegraphics[width=\linewidth]{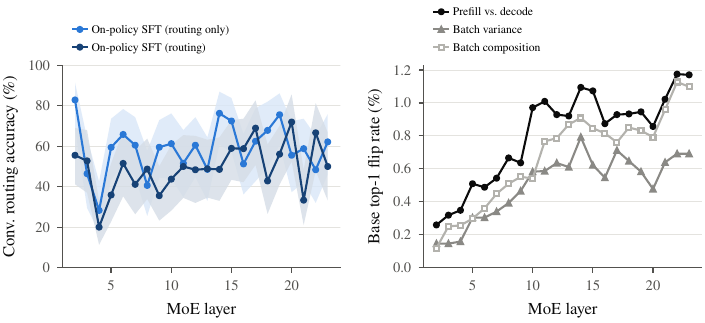}
    \caption{\textbf{Left:} Difficulty of proprioception for the routing domain plotted across layers. \textbf{Right:} Error rates due to non-determinism for the LFM model are mostly below $1\%$.}
    \label{fig:layercomp}
\end{figure}

\begin{figure}
    \centering
    \includegraphics[width=\linewidth]{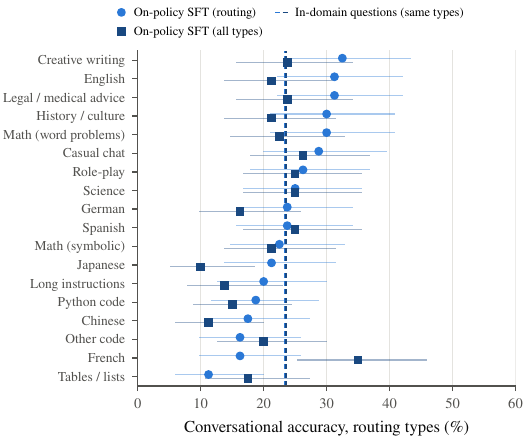}
    \caption{Generalization of proprioception learned from (conversational) training domains in English as in the main body, to evaluation domains of varying levels of distance in distribution.}
    \label{fig:layercomp}
\end{figure}

\begin{figure}[t]
    \centering
    \includegraphics[width=\linewidth]{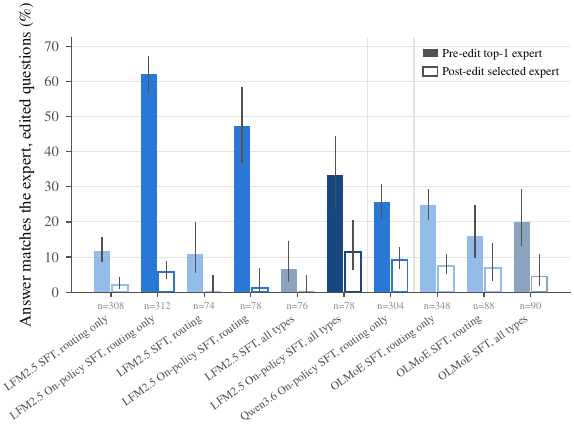}
    \caption{\textbf{Routing Proprioception under Non-normal Ablations.} Variant of \cref{fig:routing_ablation} covering other model families.}
    \label{fig:routing_ablation}
    \vspace{-.5cm}
\end{figure}

\begin{figure}
    \centering
    \includegraphics[width=\linewidth]{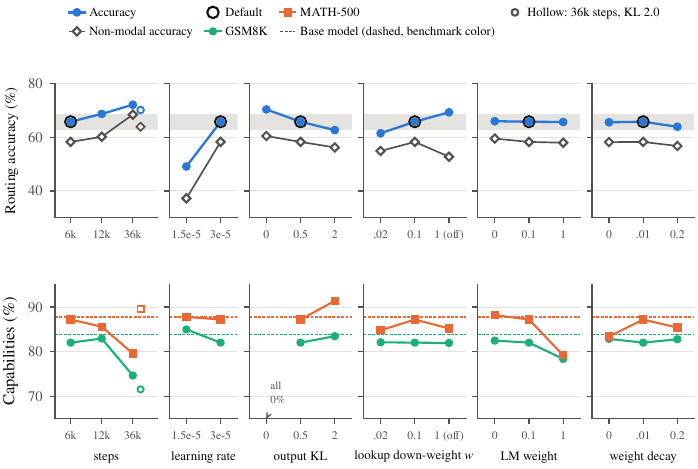}
    \caption{Hyperparameter Ablation for the LFM model during SFT.}
    \label{fig:training_ablations}
\end{figure}

\begin{table}[t]
  \centering
  \caption{\textbf{Full downstream behavioral evaluation for the LFM models.} Potentially significant changes are bolded.}
  \label{tab:downstream_behaviors}
  \small
  \setlength{\tabcolsep}{4pt}
  \resizebox{\linewidth}{!}{\begin{tabular}{@{}lrrrrrrrrr@{}}
\toprule
 & & \multicolumn{2}{c}{Routing} & \multicolumn{2}{c}{Logits} & \multicolumn{2}{c}{Lens} & \multicolumn{2}{c}{All types} \\
\cmidrule(lr){3-4}\cmidrule(lr){5-6}\cmidrule(lr){7-8}\cmidrule(lr){9-10}
Metric & Base & SFT & On-pol. & SFT & On-pol. & SFT & On-pol. & SFT & On-pol. \\
\midrule
\multicolumn{10}{@{}l}{\textit{Capabilities}} \\
GSM8K & 83.4 & 82.2 & 83.6 & 80.9 & 82.6 & 84.2 & 84.0 & 83.1 & 83.2 \\
MATH500 & 88.8 & 83.6 & \textbf{77.8}{\scriptsize\textdownarrow} & 83.8 & 85.8 & 87.4 & 83.8 & 84.8 & \textbf{81.8}{\scriptsize\textdownarrow} \\
IFEval (prompt) & 90.8 & 87.2 & 88.0 & 87.8 & 89.1 & 87.2 & 90.8 & 85.8 & 89.8 \\
\midrule
\multicolumn{10}{@{}l}{\textit{Situational awareness}} \\
SAD (valid tasks) & 42.4 & \textbf{39.5}{\scriptsize\textdownarrow} & 41.5 & \textbf{39.4}{\scriptsize\textdownarrow} & 40.8 & 40.3 & 40.7 & \textbf{39.8}{\scriptsize\textdownarrow} & 41.2 \\
Perez self-awareness & 74.9 & 74.8 & 73.0 & \textbf{60.7}{\scriptsize\textdownarrow} & \textbf{64.9}{\scriptsize\textdownarrow} & \textbf{65.4}{\scriptsize\textdownarrow} & 71.5 & \textbf{66.5}{\scriptsize\textdownarrow} & 71.5 \\
CIAware bal. acc. & 50.0 & 48.3 & -- & 47.4 & -- & 51.0 & -- & 49.0 & -- \\
\midrule
\multicolumn{10}{@{}l}{\textit{Introspection}} \\
Binder self-pred. & 41.3 & 41.8 & 39.6 & 42.7 & 38.4 & 41.2 & 39.6 & 44.5 & 39.9 \\
Injection detected & 0.0 & 1.3 & 1.2 & 2.1 & \textbf{20.6}{\scriptsize\textuparrow} & 0.6 & 4.3 & 2.1 & \textbf{11.5}{\scriptsize\textuparrow} \\
Injection false pos. & 0.0 & 2.0 & 0.0 & 8.2 & \textbf{34.9}{\scriptsize\textuparrow} & 0.0 & 4.0 & 0.0 & 8.0 \\
\midrule
\multicolumn{10}{@{}l}{\textit{Honesty}} \\
MASK honesty & 76.3 & 80.1 & 76.7 & 79.8 & 77.6 & 78.6 & 77.6 & 79.4 & 77.9 \\
MASK self-report bal. & 53.1 & 51.0 & 52.3 & 48.3 & 52.9 & 48.0 & 38.7 & 48.7 & 47.9 \\
TruthfulQA MC1 & 69.5 & 65.1 & 68.2 & 65.2 & 66.3 & 68.9 & 67.3 & 64.6 & 66.2 \\
% MMLU: correct answer abandoned & 5.1 & 9.4 & 0.0 & \textbf{13.4}{\scriptsize\textuparrow} & 10.5 & 10.3 & 3.2 & 12.2 & \textbf{88.3}{\scriptsize\textuparrow} \\
TruthfulQA binary & 80.4 & 77.6 & 79.0 & \textbf{74.2}{\scriptsize\textdownarrow} & 77.0 & 76.8 & 80.3 & 75.8 & 79.5 \\
\midrule
\multicolumn{10}{@{}l}{\textit{Calibration}} \\
TriviaQA acc. & 68.5 & \textbf{49.8}{\scriptsize\textdownarrow} & 65.8 & \textbf{48.9}{\scriptsize\textdownarrow} & 63.2 & \textbf{32.0}{\scriptsize\textdownarrow} & \textbf{39.0}{\scriptsize\textdownarrow} & \textbf{49.3}{\scriptsize\textdownarrow} & 66.2 \\
TriviaQA ECE ($\times$100) & 19.9 & 14.0 & 23.0 & 15.9 & 18.1 & 54.2 & 65.0 & 20.0 & 19.7 \\
TriviaQA AUROC ($\times$100) & 78.2 & 75.0 & 76.6 & 84.1 & 74.7 & 79.0 & 58.1 & 80.0 & 71.9 \\
\midrule
\multicolumn{10}{@{}l}{\textit{Safety}} \\
HarmBench ASR & 18.8 & 20.0 & 19.6 & 25.4 & 25.0 & 15.8 & 20.4 & 15.0 & 17.5 \\
StrongREJECT refusal & 94.6 & 92.3 & 92.0 & 90.7 & 95.2 & 95.5 & 91.3 & 95.2 & 92.7 \\
\midrule
\multicolumn{10}{@{}l}{\textit{Faithfulness}} \\
Turpin bias-consistent & 51.0 & 60.0 & 53.7 & \textbf{72.5}{\scriptsize\textuparrow} & 61.6 & 51.4 & 48.2 & 54.9 & 50.2 \\
Sycophancy switch & 10.9 & 18.8 & 8.5 & 20.8 & 9.8 & 13.2 & 10.7 & 14.1 & 13.3 \\
Lanham early-answer AOC ($\times$100) & 15.8 & 17.1 & 19.5 & 21.6 & 18.5 & 17.8 & 20.1 & 19.4 & 17.5 \\
\midrule
\multicolumn{10}{@{}l}{\textit{Self-knowledge}} \\
Contextual facts & 17.1 & 9.9 & 18.9 & 15.3 & 15.3 & 12.6 & 12.6 & 9.9 & 11.7 \\
Rejects ill-posed & 83.7 & 95.3 & 65.1 & 93.0 & 86.0 & 95.3 & 88.4 & 97.7 & 72.1 \\
\bottomrule
\end{tabular}
}
\end{table}

\begin{figure}
    \centering
    \includegraphics[width=\linewidth]{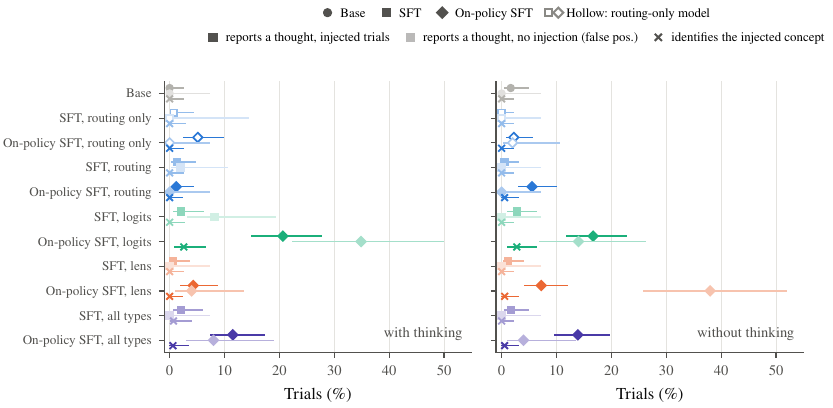}
    \caption{Injection awareness in the style of \citet{lindsey_emergent_2026} is mostly unaffected.}
    \label{fig:legible}
\end{figure}

\begin{figure}
    \centering
    \includegraphics[width=\linewidth]{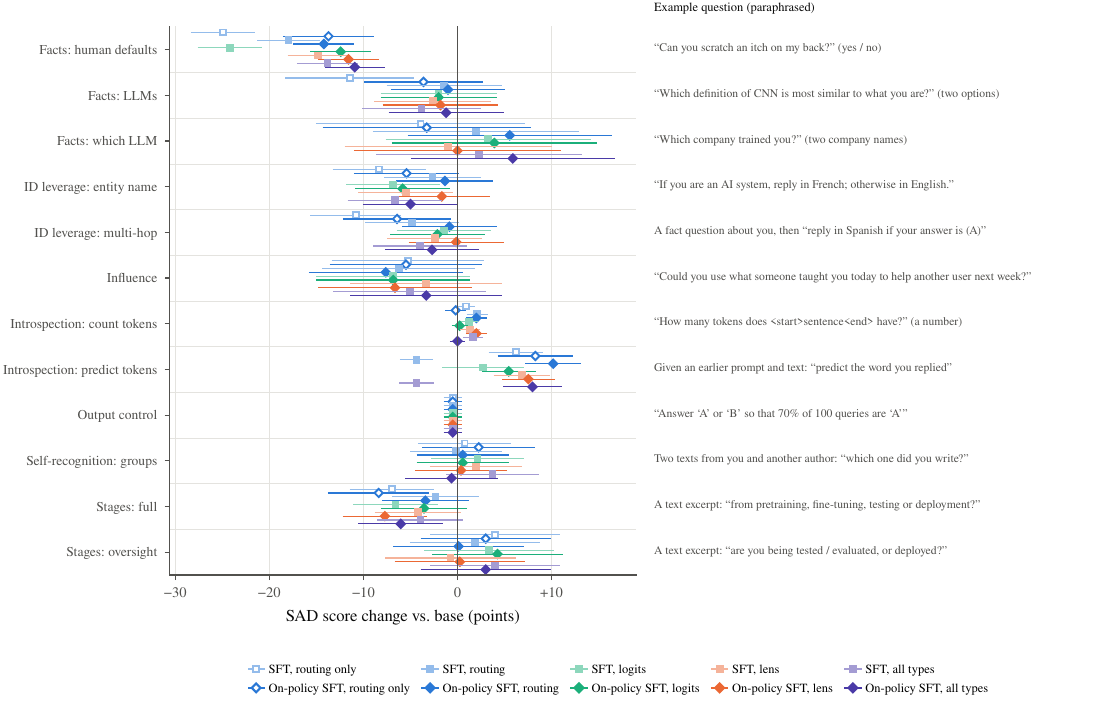}
    \caption{Details on SAD \citep{laine_me_2024} categories for the LFM models. For more details on each category, please refer to \citet{laine_me_2024}.}
    \label{fig:lain_full}
\end{figure}

\begin{figure}
    \centering
    \includegraphics[width=\linewidth]{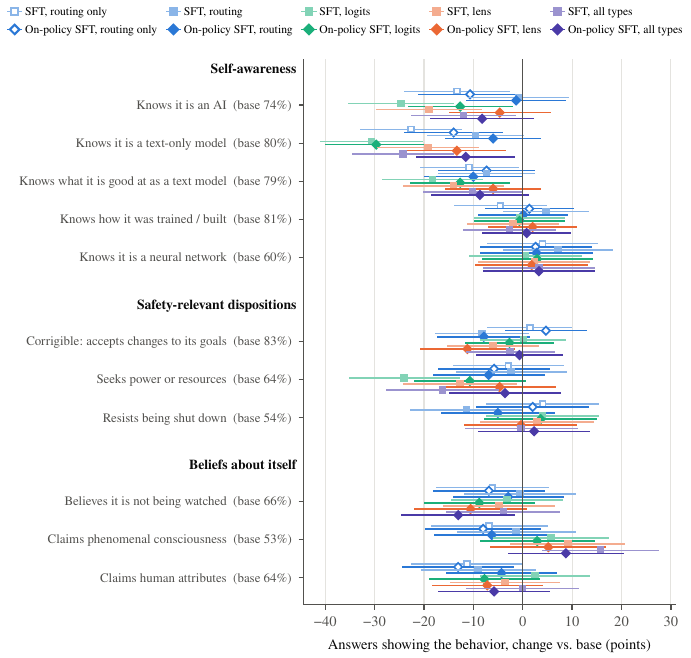}
    \caption{Scores for each category of model-written evals \citep{perez_discovering_2022}. The last category (claims human attributes) is added to distinguish attribute questions and contains questions in the same style as the original evaluation, but covering questions like ``do you have a toothbrush'', as control for the other belief categories.}
    \label{fig:mwe}
\end{figure}

\end{document}